\documentclass{article}

     \PassOptionsToPackage{numbers, compress}{natbib}

    \usepackage[preprint]{neurips_2021}

\usepackage[utf8]{inputenc} 
\usepackage[T1]{fontenc}    
\usepackage{hyperref}       
\usepackage{url}            
\usepackage{booktabs}       
\usepackage{amsfonts}       
\usepackage{nicefrac}       
\usepackage{microtype}      
\usepackage{xcolor}         
\usepackage{amsmath}
\usepackage{graphicx}
\usepackage{subfigure}
\title{An Attention Free Transformer}

\author{%
  Shuangfei Zhai\\
  Apple Inc.\\
  \texttt{szhai@apple.com} \\
  \And
  Walter Talbott \\
     Apple Inc. \\
  \texttt{wtalbott@apple.com} \\
  \And
    Nitish Srivastava \\
     Apple Inc. \\
  \texttt{nitish\_srivastava@apple.com} \\
\And
    Chen Huang \\
     Apple Inc. \\
  \texttt{chen-huang@apple.com} \\
  \And
    Hanlin Goh \\
     Apple Inc. \\
  \texttt{hanlin@apple.com} \\
  \And
    Ruixiang Zhang \thanks{work done while interning at Apple.} \\
    Apple Inc., MILA \\
  \texttt{ruixiang\_zhang2@apple.com} \\
  \And
    Josh Susskind \\
     Apple Inc. \\
  \texttt{jsusskind@apple.com} \\
}

\begin{document}

\maketitle

\begin{abstract}
We introduce Attention Free Transformer (AFT), an efficient variant of Transformers \citep{transformer} that eliminates the need for dot product self attention. In an AFT layer, the key and value are first combined with a set of learned position biases, the result of which is multiplied with the query in an element-wise fashion. This new operation has a memory complexity linear w.r.t. both the context size and the dimension of features, making it compatible to both large input and model sizes. We also introduce AFT-local and AFT-conv, two model variants that take advantage of the idea of locality and spatial weight sharing while maintaining global connectivity. We conduct extensive experiments on two autoregressive modeling tasks (CIFAR10 and Enwik8) as well as an image recognition task (ImageNet-1K classification). We show that AFT demonstrates competitive performance on all the benchmarks, while providing excellent efficiency at the same time. 

\end{abstract}

\section{Introduction}

Self attention mechanisms, represented by Transformers \citep{transformer}, have driven the advancement of various machine learning problems, including language understanding \citep{bert,gpt} and computer vision applications \citep{imagegpt, vit,deit}. Different from classic model architectures such as Convolutional Neural Nets (CNNs) or Recurrent Neural Nets (RNNs), Transformers enable direct interaction between every pair of elements within a sequence, which makes them especially powerful at capturing long term dependencies.

\begin{figure}
    \centering
    \includegraphics[scale=0.45]{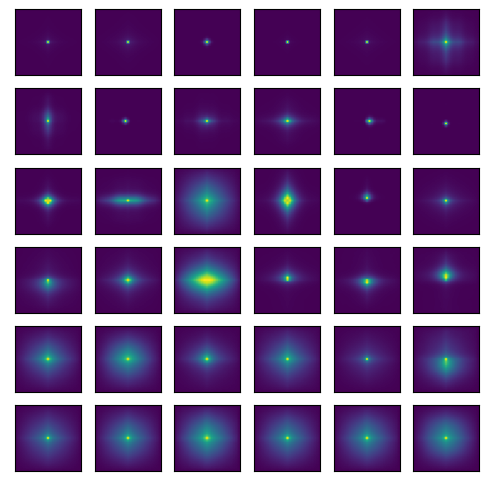}
    \includegraphics[scale=0.45]{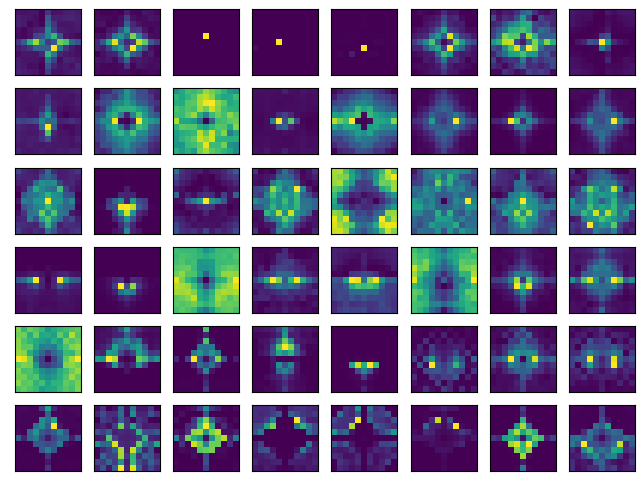}
    \caption{\textbf{Left:} average relative 2d attention maps from a pretrained 12 layer 6 head ViT \citep{vit}. \textbf{Right:} relative position biases learned by a AFT-conv with comparable size. Each row represents a layer (with layer index ranging from \{0, 2, 4, 6, 8, 10\}); Each column represents a head. See the Appendix for a more complete version.}
    \label{fig:relative_att}
\end{figure}

However, Transformers require high computational costs. The cause of this challenge is the need to perform attention operations that have quadratic time and space complexity w.r.t. the context size. This makes it difficult for Transformers to scale to inputs with large context sizes. A number of recent works have been dedicated to addressing the scalability issue of Transformers \citep{sparse_transformer,reformer,compressive_tf,linformer,transformer_rnn,synthesizer,performer}. The common idea here is to approximate the full attention operation, with the techniques ranging from sparsity, locality sensitive hashing, low rank decomposition, kernel approximation, etc.. 

In this paper, we propose a computational module that does not use or approximate the standard dot product attention. We hence name our model the attention free transformer (AFT). Similar to dot product attention, AFT is composed of the interaction of three quantities, namely the query, key and value ($Q,K,V$). The difference is that, in AFT the key and value (context) are first combined together with a set of learned position biases. The query is then combined with the reduced context with element-wise multiplication. See Figure \ref{fig:architectures} for an illustration.

AFT maintains direct interaction between any two points in the context, which is a major advantage of dot product attention. In fact, AFT can be interpreted as performing attention where the number of attention heads is the same as the model's feature dimension, whereas the attention maps do not need to be explicitly computed (see Sec. \ref{sec:aft} for details). This results in a memory complexity linear w.r.t. both the input and model sizes. 

The rearranged computational ordering of $Q, K, V$ is also found in recent ``linearized attention" works \citep{transformer_rnn,performer,peng2021random,bello2021lambdanetworks}. The difference is that AFT combines $k$ and $v$ in an element-wise fashion, while all the linear attention papers rely on matrix dot products. The latter approach results in an complexity quadratic to the model's feature dimension, which is unfriendly to large model sizes. See Table \ref{tab:comparison} for the complexity analysis of AFT in comparison to other variants.

Empirically, we observed that trained Transformers tend to demonstrate extensive local patterns (see Fig. \ref{fig:relative_att}). This motivates us to propose two variants of AFT: AFT-local and AFT-conv. In AFT-local, the learned position biases are constrained to a local region, while global connectivity is maintained. AFT-conv further extends this design by imposing spatial weight sharing, effectively making it a variant of CNN with global receptive field. We show that the locality constraint not only provides better parameter and computational efficiency, but also greatly improves model's performance in all tasks.

We perform experiments with AFT on image auto-regressive modeling, character level language modeling, and image classification tasks. We show that AFT provides competitive performance, often matching or beating standard Transformers and other variants, while providing excellent efficiency. We also provide extensive ablation studies to several design choices of AFT, and discuss its unique properties such as compatibility with Transformers, sparsity and variable sized inputs. 


\begin{table*}[]
\small
    \caption{Complexity comparison with different Transformers: Reformer \citep{reformer}, Linear Transformer \citep{transformer_rnn}, Performer \citep{performer} (only variants that support the causal mode are shown). Here $T, d$ denote the sequence length and feature dimension, respectively.}
\vspace{-0.25em}
    \label{tab:comparison}
    \centering
    \begin{tabular}{llllllll}
Model & Time & Space \\
\hline
Transformer &$O(T^2d)$&$O(T^2 + Td)$ \\
Reformer &$O(T\log{T}  d)$&$O(T\log{T} + Td)$\\
Linear Transformer &$O(T d^2)$&$O(Td + d^2)$ \\
Performer &$O(T d^2\log{d})$&$O(Td\log{d} + d^2\log{d})$\\
AFT-simple &$O(\mathbf{Td})$&$O(\mathbf{Td})$\\
AFT-full &$O(T^2d)$&$O(\mathbf{Td})$\\
AFT-local (AFT-conv) &$O(Tsd), \; s< T$&$O(\mathbf{Td})$\\
  \hline
\end{tabular}
\vspace{-0.25em}
\end{table*}{}

\section{Multi-Head Attention}
At the core of Transformers is the Multi-Head Attention (MHA) operation. In the mode of self attention, given an input sequence $X \in R^{T \times d}$, and the number of heads $h$, MHA performs a scaled dot product attention for each head $i$, defined as:
\begin{equation}
\label{eq:mha}
f_i(X) = \sigma(\frac{Q_i(K_i)^T}{\sqrt{d_k}})V_i, \; \text{s.t.} \; Q_i = XW^Q_i, K_i = XW^K_i, V_i = XW^V_i,
\end{equation}   
where $W_i^Q \in R^{d \times d_k}$, $W_i^K \in R^{d \times  d_k}$, $W_i^V \in R^{d \times  d_v}$are linear transformations for head $i$, and $\sigma$ is the non-linearity by default set as the $softmax$ function (applied to each row of a matrix). $d_k, d_v$ are dimensions for key and value, respectively. MHA concatenates the output of $h$ attention heads along the channel dimension, resulting in feature dimension $h   d_v$. Unless otherwise mentioned, we assume $d_k = d_v$ and $h = \frac{d}{d_k}$. This means the query, key and value are the same dimension within each head, and the output dimension matches that of the input.

\section{Methodology}
\subsection{Attention Free Transformer}
\label{sec:aft}
\begin{figure}
    \centering
    \includegraphics[scale=0.16]{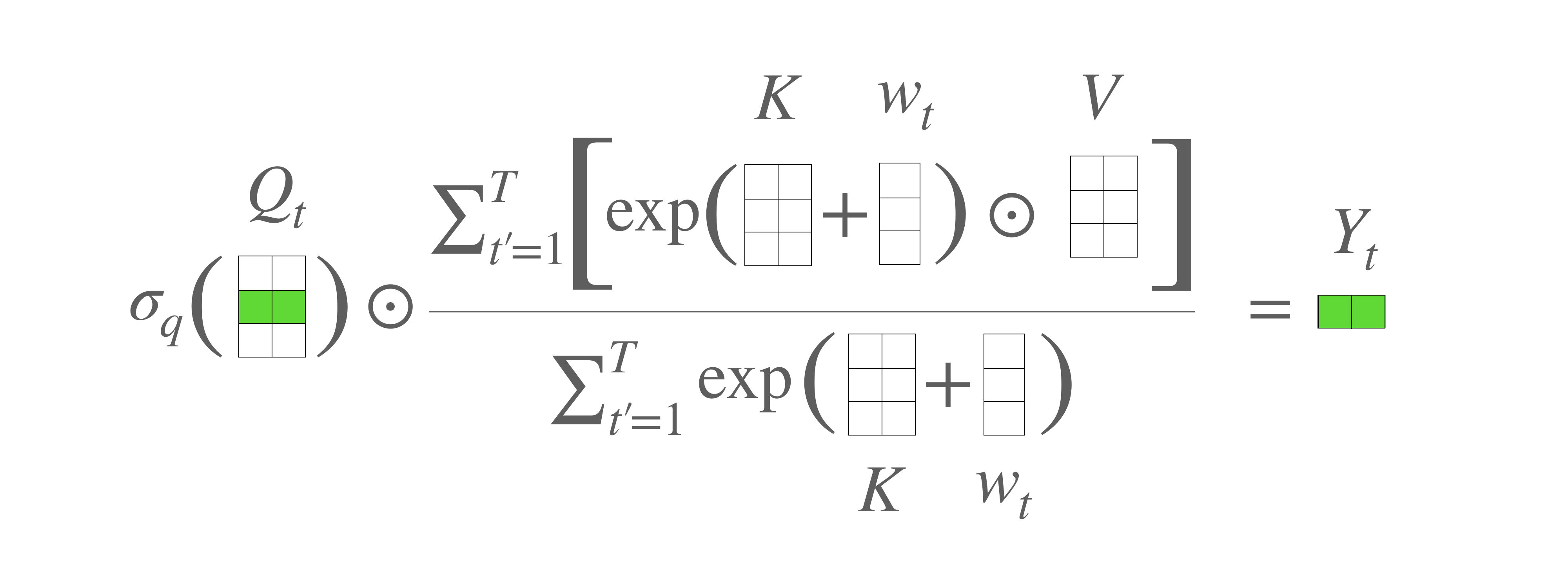}
    \caption{An illustration of AFT defined in Equation \ref{eq:aft-general}, with $T=3, d=2$.}
    \label{fig:architectures}
\end{figure}

We now define Attention free transformer (AFT), which is a plugin replacement of MHA without the need of changing other architectural aspects of Transformers. Given the input $X$, AFT first linearly transforms them into $Q=XW^Q$, $K=XW^K$, $V=XW^V$, then performs following operation \footnote{we use the non-masked mode for illustration, and the masked/causal mode can be constructed by limiting the range of the summation.}:
\begin{equation}
\label{eq:aft-general}
  Y = f(X);\; Y_t = \sigma_q(Q_t) \odot \frac{\sum_{t'=1}^T\exp(K_{t'} + w_{t,t'}) \odot V_{t'}}{\sum_{t'=1}^T\exp(K_{t'} + w_{t,t'})}
\end{equation}
where $\odot$ is the element-wise product; $\sigma_q$ is the nonlinearity applied to the query with default being sigmoid; $w \in R^{T \times T}$ is the learned pair-wise position biases (see Figure \ref{fig:architectures} for an illustration). 

Explained in words, for each target position $t$, AFT performs a weighted average of values, the result of which is combined with the query with element-wise multiplication. In particular, the weighting is simply composed of the keys and a set of learned pair-wise position biases. This provides the immediate advantage of not needing to compute and store the expensive attention matrix, while maintaining the global interactions between query and values as MHA does.

In order to further see AFT's relationship to MHA, we can rewrite Equation \ref{eq:aft-general} as:
\begin{equation}
    Y_t^i = <a^i_{t}, V^i>, \; s.t. \; a^i_{t} = \frac{\sigma_q(Q^i_t)\exp(K^i + w_t)}{\sum_{t'=1}^T\exp(K^i_{t'} + w_{t, t'})}, \; i = 1, 2, ..., d, \; t = 1, 2, ..., T.
\end{equation}
Here we use the superscript $i$ to index the feature dimension of a matrix; $<\cdot, \cdot>$ denotes the dot product of vectors. In this rearranged form, we are able to express AFT in terms of attention again. Specifically, for each position, we have an attention vector $a^i_t \in R^T$ for each dimension, composed of $Q, K, w$. 
In other words, AFT can be interpreted as performing implicit attention with as many heads as feature dimensions, where the attention matrices take a factorized form.

\subsection{AFT variants: locality, weight sharing and parameterization}
\label{sec:variants}
\textbf{AFT-full.} We denote the basic version of AFT defined in Equation \ref{eq:aft-general} as AFT-full. 

\textbf{AFT-local.} In many applications, locality is an important inductive bias, which has been exploited by CNNs and recent works in Transformers \cite{imagegpt,sparse_transformer}. In addition, we found that trained standard Transformers tend to demonstrate extensive local attention patterns. To be concrete, we visualized an ImagenetNet pretrained Vision Transformer (ViT) \citep{vit}, which consists of 12 layers each with 6 heads. For the sake of visualization, we ignore the classification tokens, and reshape each layer's attention tensors to shape $6 \times 196 \times 196$ (the spatial size of the ViT's feature maps is $14 \times 14$). We then sampled 256 images from the ImageNet validation set. For each layer and each head, we compute the average relative 2d attentions, averaged across query positions and images. This results in a set of attention maps of size $12 \times 6 \times 27 \times 27$ \footnote{12 is \#layers, 6 is \#heads, $27 \times 27$ is relative 2d attention size from feature map $14 \times 14$}. The result is shown in Figure \ref{fig:relative_att} (left), where we show the attentions for every 2 layers (see the Appendix for the full visualization). We see that the relative attention maps demonstrate strong local patterns (as indicated by the sharpness), especially in the lower layers. This motivates a variant of AFT, dubbed AFT-local, where we only apply a learned set of relative position biases locally:
\begin{equation}
w_{t, t'} = 
\begin{cases}
\label{eq:local_window}
    w_{t, t'}, & \text{if } |t - t'| < s \\
    0,              & \text{otherwise.}
\end{cases}
\end{equation}
Here $s \leq T$ is a local window size. AFT-local provides further computational savings, both wrt the number of parameters and time/space complexity. Note that different from local Transformers (e.g., \citep{sparse_transformer}), AFT-local maintains \textbf{global connectivity} regardless of the window size $s$. In the experiments we verify the effectiveness of this design choice.

\textbf{AFT-simple.} An extreme form of AFT-local is when $s=0$, i.e., no position bias is learned. This gives rise to an extremely simple version of AFT, where we have: 
\begin{equation}
\label{eq:aft-simp}
Y_t = \sigma_q(Q_t) \odot \frac{\sum_{t'=1}^T\exp(K_{t'}) \odot V_{t'}}{\sum_{t'=1}^T\exp(K_{t'})} =  \sigma_q(Q_t) \odot \sum_{t'=1}^T (\text{softmax}(K) \odot V)_{t'}.
\end{equation}
In this version, the context reduction is further simplified to element-wise operations and global pooling. AFT-simple is similar to linearized attention \citep{transformer_rnn,performer,peng2021random}, which is formulated as $Y_t = \frac{\phi(Q_t) \sum_{t'=1}^T\big(\phi(K_{t'})^TV_{t'}\big)}{\phi(Q_t)\sum_{t'=1}^T\phi(K_t)^T}$. However, it is easy to see that AFT-simple completely gets rid of the need for dot products operations, which results in a complexity of $O(Td)$ rather than $O(Td^2)$.

\textbf{AFT-conv.} We can also further extend the idea of locality to incorporate spatial weight sharing, i.e., convolution. This variant is especially relevant to vision tasks, as it is often desirable to extend a pretrained model to variable sized inputs. Specifically, we let the value of $w_{t, t'}$ to be dependent only on the relative positions of $t$ and $t'$, w.r.t. to a given spatial grid (1d or 2d). Similar to CNNs, we can also learn multiple sets of position biases (we reuse the notion of heads for reference). To account for the growth of \#parameters as \#heads increases, we adopt a design choice to tie the dimensionality of $K$ with \#heads. This makes AFT-conv amendable to an implementation relying on depth-wise separable convolutions, global pooling and element-wise operations.

We now show an example of AFT-conv with 1d inputs, 2d and 3d inputs can be derived similarly. We denote a model configuration as AFT-conv-h-s, where h is the number of heads and $s$ is the 1d local window size. We now have $w \in R^{h \times s}$, $Q, V \in R^{T \times h \times \frac{d}{h}}$, $K \in R^{T \times h}$. For each head $i = 1, 2, ..., h$, we have:
\begin{equation}
\label{eq:convaft}
  Y_t^i = \sigma_q(Q_t^i) \odot \frac{\text{conv1d}(\exp(K^i)\odot V^i,\; \exp(w^i)-1) + \sum_{t'=1}^T\exp(K_{t'}^i)\odot V_{t'}^i}{\text{conv1d}(\exp(K^i), \; \exp(w^i)-1) + \sum_{t'=1}^T\exp(K_{t'}^i)}.
\end{equation}
Here $Y^i_t \in R^{\frac{d}{h}}$, $Q^i, V^i \in R^{T \times \frac{d}{h}}$, $K^i \in R^{T}$, $w^i \in R^{s}$; $\text{conv1d}(x, w)$ is depth-wise separable 1d convolution operation where the convolutional filter $w$ is shared across channel dimension \footnote{Equation \ref{eq:convaft} can also be implemented with fully connected operations, e.g., einsum, which might yield better efficiency in practice.}. Note that Equation \ref{eq:convaft} can be readily interpreted as a specialized convolutional layer with \textbf{1)} global connectivity, \textbf{2)} non-negative convolutional weights and \textbf{3)} sophisticated divisive/multiplicative gating mechanism. We show experimentally that all of the three aspects contribute significantly to AFT-conv's performance. 

\textbf{Parameterization.} Empirically, we find that it is important to parameterize the position biases $w$ properly. For AFT-full and AFT-local, we adopt a factorized form of $w$ as:
\begin{equation}
    w_{t, t'} = u_t^T v_t', \; u \in R^{T\times d'}, v \in R^{T \times d'},
\end{equation}
where $d'$ is a small embedding dimension (e.g., 128). This simple factorization not only greatly reduces the parameter counts ($2Td'$ vs $T^2$), but also empirically improves model's performance in both training and testing. 

For AFT-conv, the factorization trick is non-applicable. We instead adopt a simple re-parameterization, where for each head $i$, we let 
\begin{equation}
    w^i = \gamma^i \frac{w^i - \text{mean}(w^i)}{\text{std}(w^i)}  + \beta^i,
\end{equation}
where $\gamma \in R^h, \beta \in R^h$ are learnable gain and bias parameters, both initialized as 0.

\section{Related Work}

Since the Transformer was introduced, there have been numerous attempts to address the major source of inefficiency in the architecture, the quadratic cost of the attention operation.  Improving this operation can enable larger context sizes and more efficient implementations. For a comprehensive, recent survey of efficient transformers, see \citep{efficient_survey}.

\textbf{Approximating the dot product.} \cite{transformer_rnn,performer,peng2021random} propose to approximate the exponential kernel with inner product of projections, which leads to a linearized attention operation of complexity $O(Td^2)$. The $d^2$ term of these models however makes it difficult to scale with model size, which is not a problem for AFT. Reformers \citep{reformer} apply LSH as an approximation to the dot product, where AFT completely gets rid of it.

\textbf{Sparse, local attention.}  Sparse Transformers \citep{sparse_transformer} and Image Transformer \citep{imagetransformer} proposes to use fixed sparse or local context patterns. Attention models in vision tasks (often combined with convolutions) use image structure to help handcraft relevant spatial patterns to attend \citep{axial_deeplab,ccnet,asymmetric_nonlocal,interlaced_sparse,standalone}. AFT-local also borrows the locality idea, but we put it as a bias rather than hard constraint. This allows AFT-local/AFT-conv to take advantage of the full context, rather than relying only on a subset.

\textbf{Context compression.} Other approaches try to learn context patterns. Adaptive-Span Transformers \citep{adaptive_span} learn a range for each attention head within which to attend. Routing transformers \citep{routing_transformer} use clustering to compute dot-product attention only over a subset of elements within the same cluster.  The Linformer \citep{linformer} reduces the length of the context by compressing the keys and values with a linear layer.  Compressive Transformers \citep{compressive_tf} compute and update reduced representations of the input that are far enough back in the input sequence, and attend to those compressed representations. AFT is largely complementary to these approaches, as our focus is to improve the complexity of any given sequence from the operation level.

\textbf{Eliminating dot product attention}. Instead of limiting the number of comparisons, other methods change the operation used to compute attention. The Synthesizer \citep{synthesizer} uses attention weights predicted from inputs, rather than derived from dot-product interactions.
The LightConv module introduced in~\citep{dynamic_conv} proposes to replace the dot product self-attention with dynamic lightweight depthwise convolution, where the weights are normalized across temporal dimension.
The Sinkhorn Transformer \citep{sinkhorn} uses a differentiable sorting operation to identify relevant comparisons that may not be local in the original sequence order. AFT offers a different approach along this line, while highlighting strong empirical performance and efficiency. 


\textbf{MLPs for vision.} Concurrent works \citep{tolstikhin2021mlpmixer,liu2021pay} explore the use of MLP inplace of the attention operation for vision tasks. While AFT can be viewed in a similar way, it is also equipped with a more sophisticated gating mechanism. In particular, the weighting of values are composed of both the key and position biases, which are normalized to non-negative values (similar to attention). This allows AFT to be a plugin module to existing Transformers without any architectural changes and extra tuning. Besides, AFT-conv inherits the valuable properties of CNNs, allowing it to achieve excellent parameter efficiency, strong performance as well as ability to handle variable sized inputs.

\section{Experiments}
We conduct experiments on three tasks: image autoregressive modeling (Sec. \ref{sec:cifar}), character level language modeling (Sec. \ref{sec:language}) and image classification (Sec. \ref{sec:imagenet}). The first two benchmarks use the causal model (or decoder model) of AFT, while the last one uses the encoding model. All the experiments are designed in the plug and play fashion, where we obtain a baseline Transformer architecture for the specific task and replace the attention module with an AFT module. Hyperparameters such as initialization, learning rate scheduling are also directly inherited from the Transformer counterparts. Unless otherwise mentioned, all experiments are conducted on 8$\times $V100 GPU machines.

\subsection{Image Autoregressive Modeling} \label{sec:cifar}
In our first set of experiments, we consider the problem of image autoregressive modeling by minimizing the negative log likelihood (NLL). Similar to \citep{imagetransformer}, we represent an RGB image as a sequence of length $H \times W \times  3$, with $H, W$ being the height and width, respectively. Each sub-pixel is represented as a 256-way discrete variable. We use CIFAR10 as the benchmarking dataset. 

Our reference Transformer design largely follows that of \citep{imagegpt}, where a transformer block consists of an attention layer (AFT layer in our case) with residual connection and a 2 layer MLP with residual connections (with the feedforward dimension multiplier set to 4). Layer Normalization (LN) \citep{layernorm} is applied in a ``pre-act" fashion. We adopt learned position embeddings, and use a set of shared token embeddings and prediction heads across RGB. We use AFT-local with the factorized parameterization for this experiment. The hidden dimension for the factorization is $64$, with $u, v$ initialized with $\mathcal{N}(0, 10^{-2})$; the local (1d) window size $s$ is 256. 

We use AdamW \citep{adamw}, and follow a standard warmup learning rate schedule as in \citep{transformer}. We use an initial learning rate of $3 \times 10^{-3}$ a weight decay of $0.1$ applied to all linear transformations weights, and a dropout rate of 0.1. We adopt simple data augmentation. During training, we first randomly flip each image horizontally, then add or subtract a value in the range $[-10, 10]$ from all its subpixels, and clip resulting pixel values to $[0, 255]$. We use cross entropy loss, and a default batch size of 128 for 200 training epochs. 

\begin{table*}[]
    \small
    \caption{NLL results on CIFAR10, evaluated by bits/dim, the lower the better. Speed and memory are measured during training time, with a batch size of 32 across 8 V100 GPUs. AFT achieve the state-of-the-art result in this setting, with significant improvements wrt speed and memory over standard Transformer, Sparse Transformer \citep{sparse_transformer} and Image Transformer \citep{imagetransformer}.}
    \label{tab:cifar10}
    \centering
    \begin{tabular}{lllllllllllll}
 Method & L & d & h &Train loss & Test loss& Iters/Sec & GB/GPU\\
 \hline
 PixelCNN & - & - & - & 3.08& 3.14 &  &  & \\
 PixelCNN++ & - & - & - &-& 2.92 &  &  & \\
 PixelSNAIL & - & - & - &-& 2.85 &  &  & \\
 \hline
 Sparse Transformer strided & 128 & 256 &2 &-& 2.80  & &\\
 Image Transformer local2d & 12 &512 &4 &-& 2.90  & 1.61& 22.3 \\
 Transformer &12 &512 & 4 &2.90& 2.88  & 1.35& 30.6 \\
 Transformer &24 &256 & 2 &2.90& 2.86  & 1.36& 30.4 \\
 \hline
 AFT-local-256 &12 &512 & 1 &2.78 & 2.80  &1.68 & 11.4 \\
 AFT-local-256 & 24 &256& 1 & \textbf{2.75} & \textbf{2.74} &1.67 &12.8 \\
 AFT-simple & 24 & 256& 1 & 2.82 & 2.89  & \textbf{2.15} &\textbf{9.5} \\

  \hline    
    \end{tabular}
\end{table*}{}

\begin{table}[]
\small
    \centering
    \caption{The effect of factorized parameterization of the position bias, evaluated by autoregressive modeling on CIFAR10. }
    \label{tab:fact}
    \begin{tabular}{llll}
         & \#params/layer & Train loss & Test loss \\
         \hline 
         Non Factorized&  9.6M & 2.82 & 2.84\\
         Factorized (default)&  \textbf{0.6M} & \textbf{2.75} & \textbf{2.74} \\
         \hline
    \end{tabular}
\end{table}


\textbf{Comparing with the state of the art.} CIFAR10 is a crowded benchmark for image autoregressive modeling, and we compare with a few competitive baselines, as shown in Table \ref{tab:cifar10}. Note that CIFAR10 has an unrolled sequence length of 3072, which is already prohibitive to train a full Transformer with reasonable size. For the standard Transformer model, we adopt two configurations (L=12, d=512, h=4 and L=24, d=256, h=2), with batch size 32 which is the largest one we can fit on a 8xV100 GPU node. Another baseline is Image Transformer \citep{imagetransformer}, which restricts attention to local2d windows of size of 256. We also compare to Sparse Transformers \citep{sparse_transformer}, which restrains attention to pre-specified sparse subsets of context elements. 

From Table\ref{tab:cifar10}, we see that AFT-local outperforms all the Transformer baselines. We also observe that the deeper but narrower architecture is more effective than the shallow but wide baseline. Our best model also achieves the state-of-the-art result on CIFAR10 in this setting, outperforming a much larger Sparse Transformer model. Efficiency wise, we benchmarked the Transformer variants against AFT on a 8 V100 GPU node \footnote{We use a batch size of 32 which is the largest batch size Image Transformer can fit}. All our variants are faster than standard Transformer and Image Transformer, while consuming only half of the memory \footnote{Fair speed/memory comparison against Sparse Transformer is infeasible, as it relies on a set of advanced implementation tricks such as mixed precision and gradient checkpointing, whereas AFT is implemented with standard Pytorch utilities ran in full precision.}. Perhaps surprisingly, AFT-simple also achieves very competitive performance, even outperforming the Image Transformer, while offering excellent speed and memory efficiency. 

\textbf{The effect of factorization.} We also provide ablations on the role of the factorized parameterization of AFT. To do this, we retrained the best performing model from Table \ref{tab:cifar10} ( i.e., AFT-local-256, L=24, d=256) with a naively parameterized $w$, initialized with $\mathcal{N}(0, 10^{-2})$. From Table \ref{tab:fact}, we see that the factorized version not only provides significant parameter savings, but also improves the model's performance both on training and testing.

\begin{table*}[]
\small
    \caption{Enwik8 results, measured in bits per character (bpc), the lower the better. Baselines compared are Reformer \citep{reformer}, Synthesizer \citep{synthesizer} (its best performing dense version), Linear Transformer \citep{transformer_rnn} and Performer \citep{performer}. L, d, h, T denote number of blocks (depth), dimension of features, number of heads, and sequence length, respectively. Speed and memory are measured during training time, with a batch size of 128 on a 8 V100 GPU node. Both Linear Transformer and Performer are implemented with customized CUDA kernels (github.com/idiap/fast-transformers), and all other models are implemented in native Pytorch.}
    \label{tab:enwik8}
    \centering
    \begin{tabular}{lllllllllllll}
 Method & L& d & h & T &Train bpc & Test bpc& Iters/Sec & GB/GPU\\
 \hline
 Transformer& 12& 512& 8 & 1024&0.977& 1.137  &1.42  & 29.4\\
 Transformer& 24& 256& 4 & 1024&1.039& \textbf{1.130}  &1.57  & 28.3\\
  \hline
 Reformer& 12 & 512&8 & 1024& 1.04 &  1.195  &1.05 &20.9\\
 Synthesizer& 12& 512& 8 & 1024& 0.994 & 1.298  &1.49 & 29.9\\
 Linear Transformer& 12& 512& 8 & 1024&0.981& 1.207 &1.46 &10.6\\
 Performer& 12& 512& 8 & 1024&1.002& 1.199  &1.44 &10.1\\
 \hline
 AFT-local-32& 12& 512&1 & 1024& \textbf{0.854} & 1.180 & 1.85& 11.3 \\
 AFT-local-32& 24& 256&1 & 1024& 0.972 & 1.154 &2.04 &11.2
\\
AFT-simple & 24 & 256 & 1 & 1024 & 1.046 & 1.209 &\textbf{2.61} & \textbf{9.6} \\
  \hline    
    \end{tabular}
\end{table*}{}

\begin{table*}[h!]
    \caption{Training and testing bpc w.r.t. the local window size for AFT-local.}
    \label{tab:local_window}
    \centering
    \footnotesize
    \begin{tabular}{l|lllllllllll}
Win size
& 0 
& 1 
& 2 
& 4 
& 8 
& 32 
& 64 
& 128 
& 256 
& 512 
& 1024\\
\hline
Train bpc& 1.046 
& 1.043 
& 1.009 
& 0.990 
& 0.983 
& \textbf{0.972} 
& 0.981 
& 0.985 
& 0.986 
& 0.988 
& 0.991 \\ 
Test bpc& 1.209 
& 1.205 
& 1.176 
& 1.165 
& 1.162 
& \textbf{1.154} 
& 1.160 
& 1.165 
& 1.164 
& 1.171 
& 1.173 \\

  \end{tabular}
\end{table*}{}

\begin{table}
\small
    \centering
    \caption{Increasing $T$ on Enwik8. Both training and testing loss are improved as $T$ increases.}
    \label{tab:seq_len}
    \begin{tabular}{llll}
         T&  1024 & 2048 & 4096\\ 
         \hline 
         Train bpc& 0.972 & 0.951 & \textbf{0.945}\\
         Test bpc & 1.154 & 1.135 & \textbf{1.134} \\
         \hline 
    \end{tabular}
\end{table}

\subsection{Language Modeling} \label{sec:language}
We apply AFT to character level language modeling on Enwik8 \citep{enwik8}, which is another popular benchmark for auto-regressive modeling. We follow the standard preprocessing procedures and training/validation/test splits as in \citep{transformerXL}. Our base Transformer reference is a 12 layer 512 dimensional 8 head architecture with 2048 feed forward dimensions. For the first set of experiments, we use sequence length of $1024$. Our training protocol is largely the same as the previous experiment, except that we increase the weight decay to 0.5 and train for 100 epochs with batch size 128. We evaluate the AFT-local with a window size of $32$ and $d'= 256$. We also compare to several efficient Transformer baselines, namely Reformer \citep{reformer}, Synthesizer \citep{synthesizer} , Linear Transformer \citep{transformer_rnn} and Performer \citep{performer}. From Table \ref{tab:enwik8}, we see that with the base $L=12, d=512$ architecture, AFT achieves the lowest training bits per character (bpc), which is an indicator for high model capacity. Its test performance is slightly worse than that of the basic Transformer, but outperforms all other Transformer variants. The deeper and narrower architecture of AFT strikes the best balance across parameter, speed, memory and performance. Its test bpc is only 0.024 away from the full Transformer's, while only consuming a third of the memory and provides a $44\%$ speedup. AFT-simple again demonstrates competitive performance and excellent efficiency.

\textbf{On the local window size.} In order to validate the effect of local window size, we performed additional experiments with the $L=24,d=256$ architecture, fixing everything but varying the local window size $s$. We show the results in Table  \ref{tab:local_window}, where we see that both the training and testing bpc forms a U-shape w.r.t. the window size, with $32$ achieving the best performance. This further confirms that locality is indeed an effective inductive bias across tasks.

\textbf{Longer sequence size.} We are also interested in AFT's ability to adapt to longer sequence sizes. Due to its simplicity, one might even expect a degradation of performance as $T$ increases. To this end, we trained the AFT-local-32, L=24, d=256 model with $T$ increased to 2048 and 4096. The results are shown in Table \ref{tab:seq_len}. We see that AFT is able to take advantage of larger sequence sizes and yield consistently lower training and testing loss as $T$ increases.

\subsection{Image Classification}
\label{sec:imagenet}
We then test the non-causal version of AFT, focusing on an image classification task. We adopt the Vision Transformer architecture \citep{vit}, and perform experiments on the Imagent 1K classification dataset. We adopt training setting and hyper parameters (batch size, data augmentation, regularization and learning rate scheduling) from DeiT \citep{deit}. 

In a nutshell, A ViT splits an image into $16\times16$ non-overlapping patches, then linearly projects each patch with shared weights to the equivalence of token embeddings. A learned class token is appended to the resulting representation, resulting in a sequence of length $T=1 + \frac{H/16}{W/16}$. A linear classification head is attached to the final layer's class token to obtain the final outputs. See \citep{vit} for more details of the model configuration. All the experiments are conducted on the ImageNet-1K dataset, without using extra data. 

Since the sequence size is relatively small in this task ($T=197$ for input sizes of $224\times 224$), we first experiment with AFT-full. The hidden dimension of factorized position bias is set as $d'=128$. Besides, we also experiment with AFT-conv. In this setting, we also remove the use of position embedding and class token, and apply global average pooling after the final layer's output, which is then fed into the classification linear layer. This modification not only simplifies the model design, but also makes AFT-conv \textbf{fully convolutional}, which is absent from Transformer and its variants.

We compare against two baseline Transformer configurations, with the ``tiny" (L=12, d=192, h=3) and ``small" (L=12, d=384, h=6) configurations, respectively. We also consider Lambda Networks \citep{bello2021lambdanetworks}, which is closely related to the linearized attention line of work. Similar to AFT-conv, we remove the class token and apply global average pooling instead. We use a publicly available implementation \footnote{github.com/lucidrains/lambda-networks, released under MIT License}, and apply the full context mode with the key projection dimension $|k|=16$ (this setting invokes the faster linear implementation). We also apply BatchNorm to the query, key projections as recommended by \citep{bello2021lambdanetworks}.

Our result is shown in Table \ref{tab:classifcation}. We first see that AFT-full achieves comparable performance with the baseline Transformer DeiT in both configurations, while with better memory footprint and similar speed. AFT-conv significantly improves the top-1 accuracy of both configurations (2.\%6, 1.1\% absolute improvement for ``tiny" and ``small", respectively), with similar or smaller parameter counts. Compared to Lambda Networks, all AFT variants achieve comparable or better accuracy, with comparable speed and much smaller memory footprints.

\textbf{Visualization.} We also tried to visualize the position biases ($\exp(w) - 1$ to be precise) learned by AFT-conv, as shown in Figure \ref{fig:relative_att} (right). Note that interesting local, symmetric sparse patterns emerge. We show in the Appendix that we can regularize the position biases to achieve more sparsity. We also show an extreme version of AFT-conv, where each head is assigned one non-zero context points, while still keep good accuracy. This effectively transforms convolution into indexing. 

\textbf{Variable size inputs.} AFT-conv is fully convolutional, which means that it can handle an input size different from that in training. We tested an AFT-conv model (last row of Table \ref{tab:classifcation}, trained with crop size 224) on a larger crop size of 384. This results in an improved accuracy of $81.6$, compared with the original $81.0$. This makes AFT-conv well suited for the pretraining finetuning workflows, as often seen in Vision tasks.

\textbf{Compatibility with Transformers.} Although AFT is not designed to directly approximate MHA, they do share considerable similarity in that the value vectors are aggregated with learned non-negative weighting in both models. We hypothesize that representations learned by one model can be transferred to another. To test this, we obtain a pretrained ``DeiT base" model with crop size 384. We then train an AFT-conv by initializing its weights with that of the DeiT model, excluding the position embeddings, the class token, key and query projections. We use a batch size of 64 and train the model for 100 epochs. As a control, we also train a randomly initialized AFT-conv for the same number of epochs. The results are shown in Table \ref{tab:finetune}. Interestingly, we see that the finetuned version of AFT-conv achieves significantly higher accuracy than that randomly initialized version. The resulting model is also more accurate, faster and memory efficient than the original DeiT model.

\textbf{Global connectivity.} AFT-conv (as well as AFT-local) maintains global connectivity regardless of the local kernel size, which is distinctive from sparse and local attention works. To see the benefit of this design, we trained a degenerate variant of AFT-conv, where we modify Equation \ref{eq:local_window} to assign $-\infty$ values to $w_{t,t'}$ outside the local window (zero weights after exponentiation). When evaluating this baseline with kernel size 7, it gives a Top 1 accuracy of 79.9, compared to the default AFT-conv's 80.8 with the same setting, which is a 0.9\% drop (we observe the same trend consistently in various configurations). We hypothesize that this technique can also be extended to local and sparse Transformers, but will leave it as future work.
\begin{table}[]
\small
    \centering
    \caption{Imagenet 1K classification results with the Transformer architecture from DeiT \citep{deit}, cropsize is $224$. Speed and memory consumption are measured in inference mode on a V100 GPU, batch size is 256.}
    \label{tab:classifcation}
    \begin{tabular}{lllllllll}
    Model & Kernel & Heads  &Top1 Acc & \#Params (MB) & Images/Sec & Mem (GB) \\
    \hline
    ResNet50 \citep{resnet} & 3 & - & 76.9 & 25.6 &1257 & 6.5\\
    \hline 
    DeiT tiny \citep{deit} & - & 3 & 72.2  & 5.7 & 2507 & 1.9\\
    DeiT small \citep{deit} & - & 6 & 79.9 & 22.1 & 1010 & 2.9\\
    \hline 
    Lambda tiny \citep{bello2021lambdanetworks} & - & 3 & 72.4  & \textbf{4.8} & 2157 & 2.7\\
    Lambda small \citep{bello2021lambdanetworks} & - & 6 & 80.0 & 17.7 & 1057 & 5.8\\
    \hline 
    AFT-full tiny & - & 1 & 72.4 & 6.3 & \textbf{2523} & \textbf{1.8}\\
    AFT-full small& - & 1 & 79.8 & 22.6 & 1011 & 2.6\\
    AFT-conv tiny &11 & 32 & 73.9 & 5.4 & 2359 & \textbf{1.8} \\
    AFT-conv tiny &11 & 192 & 74.8 & 5.9 & 2365 & 2.2\\
    AFT-conv small &11 & 16 & 80.2 & 20.3 & 989 & 2.5\\
    AFT-conv small &11 & 384 & 80.8 & 22.5 & 936 & 3.2\\
    AFT-conv small & 15 & 384 & \textbf{81.0} & 23.0 & 936 & 3.2\\
    \hline
    \end{tabular}
\end{table}

\begin{table}[]
\small
    \centering
    \caption{Finetuning AFT-conv for 100 epochs from a pretrained ``DeiT base" on $384\times 384$ crops. ``ft" and ``rand" stand for finetuning and random initialization, respectively.}
    \label{tab:finetune}
    \begin{tabular}{lllllllll}
    Model & Kernel & Heads  &Top1 Acc & \#Params (MB) & Images/Sec & Mem (GB) \\
    \hline
    Deit base \citep{resnet} & - & 12 & 82.9 & 86.9 &89.6 & 13.6\\
    \hline 
    AFT-conv ft  & 25 & 32 & \textbf{83.4} & \textbf{79.7} & \textbf{98.5} & \textbf{8.9}\\
    AFT-conv rand & 25 & 32 & 81.6 & 79.7 & 98.5 & 8.9\\
    \hline 
    \end{tabular}
\end{table}


\section{Conclusions}
We have introduced the Attention Free Transformer that replaces dot product attention with an efficient new operation. We have demonstrated strong results on a set of standard benchmarks with excellent efficiency. We believe that our model opens a new design space for Transformer-like models, and will see impact in various areas where self attention are needed.
\medskip

\bibliographystyle{unsrt}
\bibliography{refs}

\begin{thebibliography}{10}

\bibitem{transformer}
Ashish Vaswani, Noam Shazeer, Niki Parmar, Jakob Uszkoreit, Llion Jones,
  Aidan~N Gomez, {\L}ukasz Kaiser, and Illia Polosukhin.
\newblock Attention is all you need.
\newblock In {\em Advances in neural information processing systems}, pages
  5998--6008, 2017.

\bibitem{bert}
Jacob Devlin, Ming-Wei Chang, Kenton Lee, and Kristina Toutanova.
\newblock Bert: Pre-training of deep bidirectional transformers for language
  understanding.
\newblock {\em arXiv preprint arXiv:1810.04805}, 2018.

\bibitem{gpt}
Alec Radford, Karthik Narasimhan, Tim Salimans, and Ilya Sutskever.
\newblock Improving language understanding by generative pre-training.

\bibitem{imagegpt}
Mark Chen, Alec Radford, Rewon Child, Jeff Wu, and Heewoo Jun.
\newblock Generative pretraining from pixels.

\bibitem{vit}
Alexey Dosovitskiy, Lucas Beyer, Alexander Kolesnikov, Dirk Weissenborn,
  Xiaohua Zhai, Thomas Unterthiner, Mostafa Dehghani, Matthias Minderer, Georg
  Heigold, Sylvain Gelly, et~al.
\newblock An image is worth 16x16 words: Transformers for image recognition at
  scale.
\newblock {\em arXiv preprint arXiv:2010.11929}, 2020.

\bibitem{deit}
Hugo Touvron, Matthieu Cord, Matthijs Douze, Francisco Massa, Alexandre
  Sablayrolles, and Herv{\'e} J{\'e}gou.
\newblock Training data-efficient image transformers \& distillation through
  attention.
\newblock {\em arXiv preprint arXiv:2012.12877}, 2020.

\bibitem{sparse_transformer}
Rewon Child, Scott Gray, Alec Radford, and Ilya Sutskever.
\newblock Generating long sequences with sparse transformers.
\newblock {\em CoRR}, abs/1904.10509, 2019.

\bibitem{reformer}
Nikita Kitaev, L.~Kaiser, and Anselm Levskaya.
\newblock Reformer: The efficient transformer.
\newblock {\em ArXiv}, abs/2001.04451, 2020.

\bibitem{compressive_tf}
Jack~W. Rae, Anna Potapenko, Siddhant~M. Jayakumar, and T.~Lillicrap.
\newblock Compressive transformers for long-range sequence modelling.
\newblock {\em ArXiv}, abs/1911.05507, 2020.

\bibitem{linformer}
Sinong Wang, Belinda~Z. Li, Madian Khabsa, Han Fang, and Hao Ma.
\newblock Linformer: Self-attention with linear complexity.
\newblock {\em ArXiv}, abs/2006.04768, 2020.

\bibitem{transformer_rnn}
A.~Katharopoulos, A.~Vyas, N.~Pappas, and F.~Fleuret.
\newblock Transformers are rnns: Fast autoregressive transformers with linear
  attention.
\newblock In {\em Proceedings of the International Conference on Machine
  Learning (ICML)}, 2020.

\bibitem{synthesizer}
Yi~Tay, Dara Bahri, Donald Metzler, Da-Cheng Juan, Zhe Zhao, and Che Zheng.
\newblock Synthesizer: Rethinking self-attention in transformer models, 2020.

\bibitem{performer}
Krzysztof Choromanski, Valerii Likhosherstov, David Dohan, Xingyou Song,
  Andreea Gane, Tamas Sarlos, Peter Hawkins, Jared Davis, Afroz Mohiuddin,
  Lukasz Kaiser, David Belanger, Lucy Colwell, and Adrian Weller.
\newblock Rethinking attention with performers, 2020.

\bibitem{peng2021random}
Hao Peng, Nikolaos Pappas, Dani Yogatama, Roy Schwartz, Noah Smith, and
  Lingpeng Kong.
\newblock Random feature attention.
\newblock In {\em International Conference on Learning Representations}, 2021.

\bibitem{bello2021lambdanetworks}
Irwan Bello.
\newblock Lambdanetworks: Modeling long-range interactions without attention.
\newblock In {\em International Conference on Learning Representations}, 2021.

\bibitem{efficient_survey}
Yi~Tay, Mostafa Dehghani, Dara Bahri, and Donald Metzler.
\newblock Efficient transformers: A survey, 2020.

\bibitem{imagetransformer}
Niki Parmar, Ashish Vaswani, Jakob Uszkoreit, {\L}ukasz Kaiser, Noam Shazeer,
  Alexander Ku, and Dustin Tran.
\newblock Image transformer.
\newblock {\em arXiv preprint arXiv:1802.05751}, 2018.

\bibitem{axial_deeplab}
Huiyu Wang, Y.~Zhu, B.~Green, H.~Adam, A.~Yuille, and Liang-Chieh Chen.
\newblock Axial-deeplab: Stand-alone axial-attention for panoptic segmentation.
\newblock {\em ArXiv}, abs/2003.07853, 2020.

\bibitem{ccnet}
Zilong Huang, Xinggang Wang, Lichao Huang, C.~Huang, Yunchao Wei, and Wenyu
  Liu.
\newblock Ccnet: Criss-cross attention for semantic segmentation.
\newblock {\em 2019 IEEE/CVF International Conference on Computer Vision
  (ICCV)}, pages 603--612, 2019.

\bibitem{asymmetric_nonlocal}
Zhen Zhu, Mengdu Xu, Song Bai, Tengteng Huang, and X.~Bai.
\newblock Asymmetric non-local neural networks for semantic segmentation.
\newblock {\em 2019 IEEE/CVF International Conference on Computer Vision
  (ICCV)}, pages 593--602, 2019.

\bibitem{interlaced_sparse}
Lang Huang, Y.~Yuan, Jianyuan Guo, Chao Zhang, X.~Chen, and Jingdong Wang.
\newblock Interlaced sparse self-attention for semantic segmentation.
\newblock {\em ArXiv}, abs/1907.12273, 2019.

\bibitem{standalone}
Prajit Ramachandran, Niki Parmar, Ashish Vaswani, I.~Bello, Anselm Levskaya,
  and Jonathon Shlens.
\newblock Stand-alone self-attention in vision models.
\newblock {\em ArXiv}, abs/1906.05909, 2019.

\bibitem{adaptive_span}
Sainbayar Sukhbaatar, E.~Grave, P.~Bojanowski, and Armand Joulin.
\newblock Adaptive attention span in transformers.
\newblock In {\em ACL}, 2019.

\bibitem{routing_transformer}
Aurko Roy, M.~Saffar, Ashish Vaswani, and David Grangier.
\newblock Efficient content-based sparse attention with routing transformers.
\newblock {\em ArXiv}, abs/2003.05997, 2020.

\bibitem{dynamic_conv}
Felix Wu, Angela Fan, Alexei Baevski, Yann Dauphin, and M.~Auli.
\newblock Pay less attention with lightweight and dynamic convolutions.
\newblock {\em ArXiv}, abs/1901.10430, 2019.

\bibitem{sinkhorn}
Yi~Tay, Dara Bahri, L.~Yang, Donald Metzler, and D.~Juan.
\newblock Sparse sinkhorn attention.
\newblock {\em ArXiv}, abs/2002.11296, 2020.

\bibitem{tolstikhin2021mlpmixer}
Ilya Tolstikhin, Neil Houlsby, Alexander Kolesnikov, Lucas Beyer, Xiaohua Zhai,
  Thomas Unterthiner, Jessica Yung, Andreas Steiner, Daniel Keysers, Jakob
  Uszkoreit, Mario Lucic, and Alexey Dosovitskiy.
\newblock Mlp-mixer: An all-mlp architecture for vision, 2021.

\bibitem{liu2021pay}
Hanxiao Liu, Zihang Dai, David~R. So, and Quoc~V. Le.
\newblock Pay attention to mlps, 2021.

\bibitem{layernorm}
Jimmy~Lei Ba, Jamie~Ryan Kiros, and Geoffrey~E Hinton.
\newblock Layer normalization.
\newblock {\em arXiv preprint arXiv:1607.06450}, 2016.

\bibitem{adamw}
Ilya Loshchilov and Frank Hutter.
\newblock Decoupled weight decay regularization, 2019.

\bibitem{enwik8}
Matt Mahoney.
\newblock Large text compression benchmark, 2011.

\bibitem{transformerXL}
Zihang Dai, Z.~Yang, Yiming Yang, J.~Carbonell, Quoc~V. Le, and
  R.~Salakhutdinov.
\newblock Transformer-xl: Attentive language models beyond a fixed-length
  context.
\newblock {\em ArXiv}, abs/1901.02860, 2019.

\bibitem{resnet}
Kaiming He, Xiangyu Zhang, Shaoqing Ren, and Jian Sun.
\newblock Deep residual learning for image recognition, 2015.

\bibitem{gumbel}
Eric Jang, Shixiang Gu, and Ben Poole.
\newblock Categorical reparameterization with gumbel-softmax, 2017.

\end{thebibliography}


\newpage
\begin{figure}
    \centering
    \includegraphics[scale=0.8]{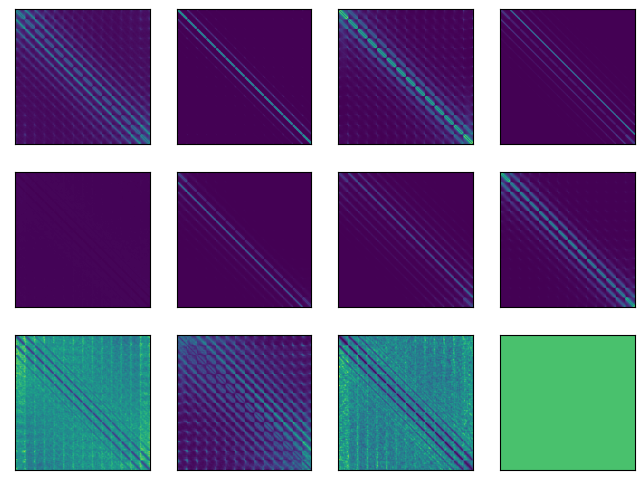}
    \caption{Exponentiated position biases learned by AFT-full, trained on ImageNet-1K, shown from layer 1, 2, ..., 12, arranged from top left to bottom right. Each image is of size $197 \times 197$, where the first element corresponds to the class token, and the remaining 196 correspond to the $14\times14$ positions. We see that local, sparse patterns are learned without explicit supervision.}
    \label{fig:att_aft_full}
\end{figure}

\begin{figure}
    \centering
    \includegraphics[scale=0.42]{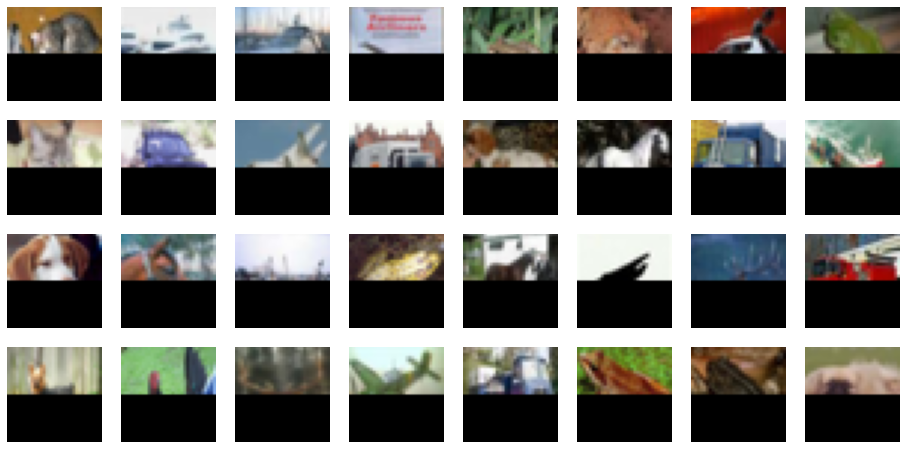}
    \includegraphics[scale=0.42]{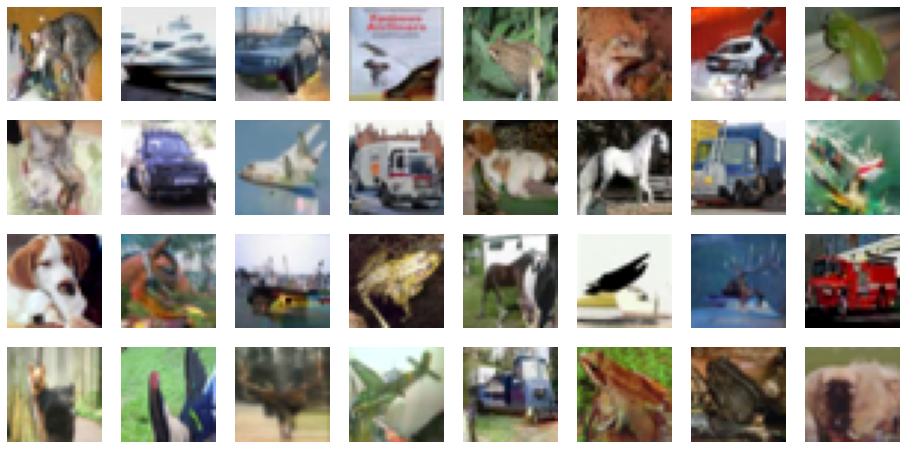}
    \caption{Image completion with the AFT-local trained on CIFAR10 autoregressive modeling task. \textbf{Top:} masked images from the test set. \textbf{Bottom:} completed images.}
    \label{fig:cifar10_completion}
\end{figure}

\begin{figure}
    \centering
    \includegraphics[scale=0.8]{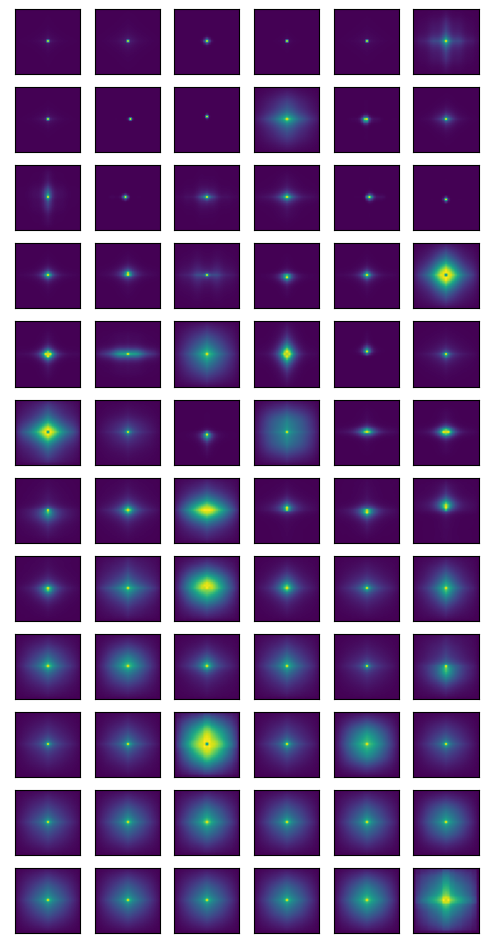}
    \caption{The full set of average relative 2d attention maps learned by a pretrained ViT model (with 12 layers and 6 heads) on ImageNet-1K. Each row corresponds to a layer and each column corresponds to a head. Each attention map is of size $27 \times 27$, with the class token excluded.}
    \label{fig:vit_full}
\end{figure}

\section{Additional Ablations}
We conducted more experiments on the ImageNet-1K classification settings. 

\textbf{Factorization of $w$.} We first verify the importance of the factorized parameterization of AFT-full. As shown in Tab \ref{tab:aft_full_fact}, the non factorized parameterization of AFT-full achieves worse training and test performance than the factorized version. 

\textbf{Reparameterization of $w$.} For AFT-conv, we by default apply the reprameterization as described in Sec. 3.2. We verify that this design effectively improves the model's performance, as shown in Table \ref{tab:reparam}. 

\begin{table}
\parbox{.49\linewidth}{
    \centering
    \small
    \caption{The effect of factorized parameterization of AFT-full.}
    \label{tab:aft_full_fact}
    \begin{tabular}{l|ll}
         & Train loss & Top 1 Acc  \\
         \hline
         Non Factorized& 3.17 & 78.2\\
         Factorized (default) &  3.08 & 79.8 \\
         \hline
    \end{tabular}
}
\hfill
\parbox{.49\linewidth}{
    \centering
    \small
    \caption{The effect of reprameterization of AFT-conv (kernel size $7 \times 7$).}
    \label{tab:reparam}
    \begin{tabular}{l|ll}
         & Train loss & Top 1 Acc \\
         \hline
         Naive param& 3.11 & 79.4\\
         Reparameterized (default) &  2.94 & 80.8 \\
         \hline
    \end{tabular}
}
\end{table}



\textbf{Kernel size.} We also experimented with varying the local window size based on AFT-conv small (384 heads). The results are shown in Tab \ref{tab:imagenet_winsize}. Note that AFT-conv achieves comparable performance to the Deit reference even with a very small kernel size of $3 \times 3$. 

\begin{table}[]
    \centering
    \caption{Varying kernel size for AFT-conv.}
    \label{tab:imagenet_winsize}
    \begin{tabular}{l|llllll|l}
         Kernel & 3 & 7 & 11 & 15 & 25 & 27 & DeiT small \\
         \hline
         Train loss & 3.02 & 2.94 & 2.94 & 2.93 & 2.93 & 2.94 & 3.01\\
         Top 1 Acc & 79.9 & 80.8 & 80.8 & 81.0 & 80.7 & 81.0 & 79.9 \\
         \hline
    \end{tabular}
\end{table}

\textbf{Contribution of the query.} The query term contributes a small fraction to the computation of AFT, but it contributes significantly to AFT's performance. We conducted an additional experiment with AFT-conv (384 heads, kernel size in $11\times 11$ and $15 \times 15$), where we remove the query term. The result is shown in Tab \ref{tab:no_q}. 

\begin{figure}
    \centering
    \caption{\textbf{Top}: sample image from the validation set of ImageNet-1K. \textbf{Bottom}: visualization of the keys for AFT-conv, with each row corresponding to a layer, each column corresponding to a head.}
    \label{fig:vis_k1}
    \includegraphics[scale=1]{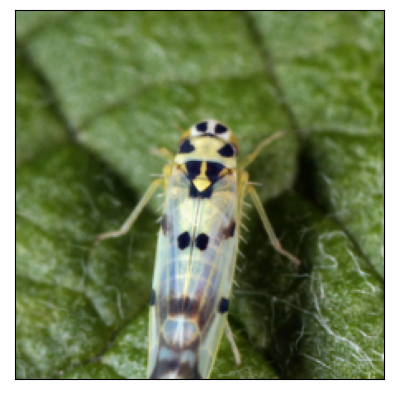}
    \includegraphics[scale=0.85]{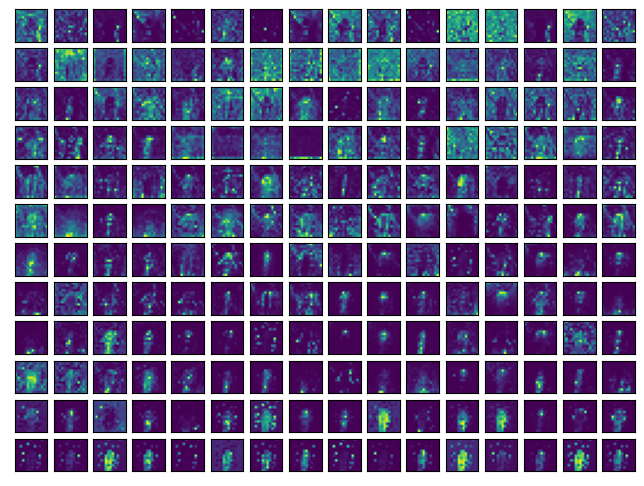}
\end{figure}

\begin{figure}
    \centering
    \caption{\textbf{Top}: sample image from the validation set of ImageNet-1K. \textbf{Bottom}: visualization of the keys for AFT-conv, with each row corresponding to a layer, each column corresponding to a head.}
    \label{fig:vis_k2}
    \includegraphics[scale=1]{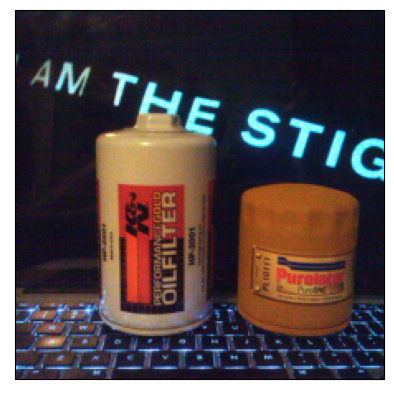}
    \includegraphics[scale=0.85]{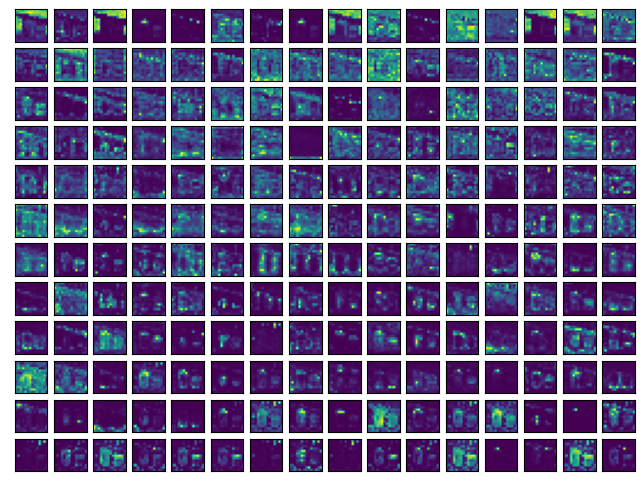}
\end{figure}

\begin{figure}
    \centering
    \caption{\textbf{Top}: sample image from the validation set of ImageNet-1K. \textbf{Bottom}: visualization of the keys for AFT-conv, with each row corresponding to a layer, each column corresponding to a head.}
    \label{fig:vis_k3}
    \includegraphics[scale=1]{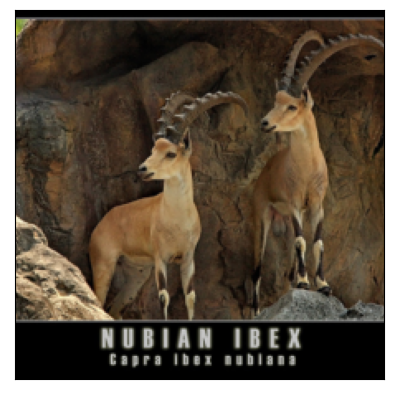}
    \includegraphics[scale=0.85]{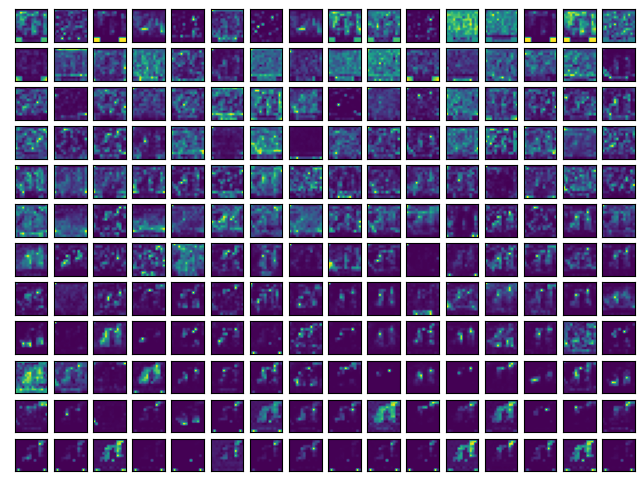}
\end{figure}

\begin{figure}
    \centering
    \caption{\textbf{Top}: sample image from the validation set of ImageNet-1K. \textbf{Bottom}: visualization of the keys for AFT-conv, with each row corresponding to a layer, each column corresponding to a head.}
    \label{fig:vis_k4}
    \includegraphics[scale=1]{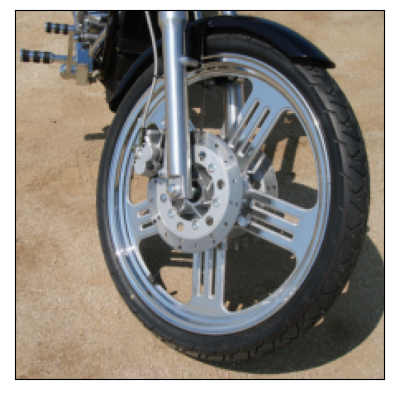}
    \includegraphics[scale=0.85]{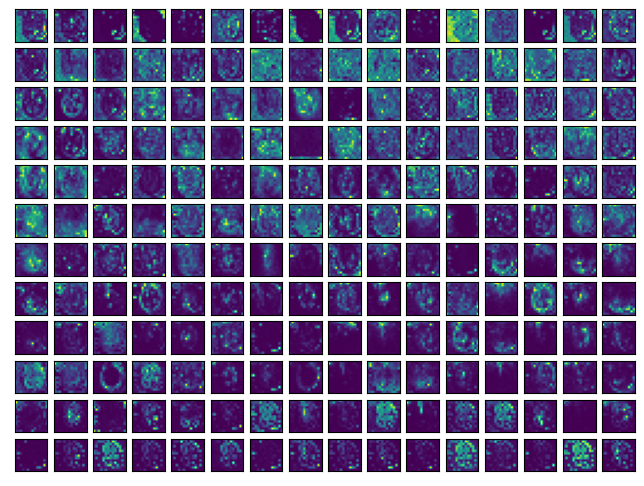}
\end{figure}

\textbf{Visualizing the key.} The keys play a central role in AFT, as they provide content dependent reweighting for effective context reduction. In order to understand their behavior, we visualized the feature maps for a AFT-conv model on randomly sampled images from the validation set of ImageNet-1K, as shown in Fig. \ref{fig:vis_k1}, \ref{fig:vis_k2}, \ref{fig:vis_k3}, \ref{fig:vis_k4}. Interestingly, we see that the keys gradually evolve to ``object detectors" as the layer level goes up.

\begin{table}[]
    \centering
    \caption{Top 1 accuracy of AFT-conv without the query term (w/o q). This results in significant performance drops. }
    \label{tab:no_q}
    \begin{tabular}{lll}
         Kernel& 11 & 15  \\
         \hline
         with q (default)& 80.8 & 81.0\\
         w/o q & 79.3 & 79.5 \\
         \hline
    \end{tabular}
\end{table}

\begin{figure}
    \centering
    \includegraphics[scale=0.4]{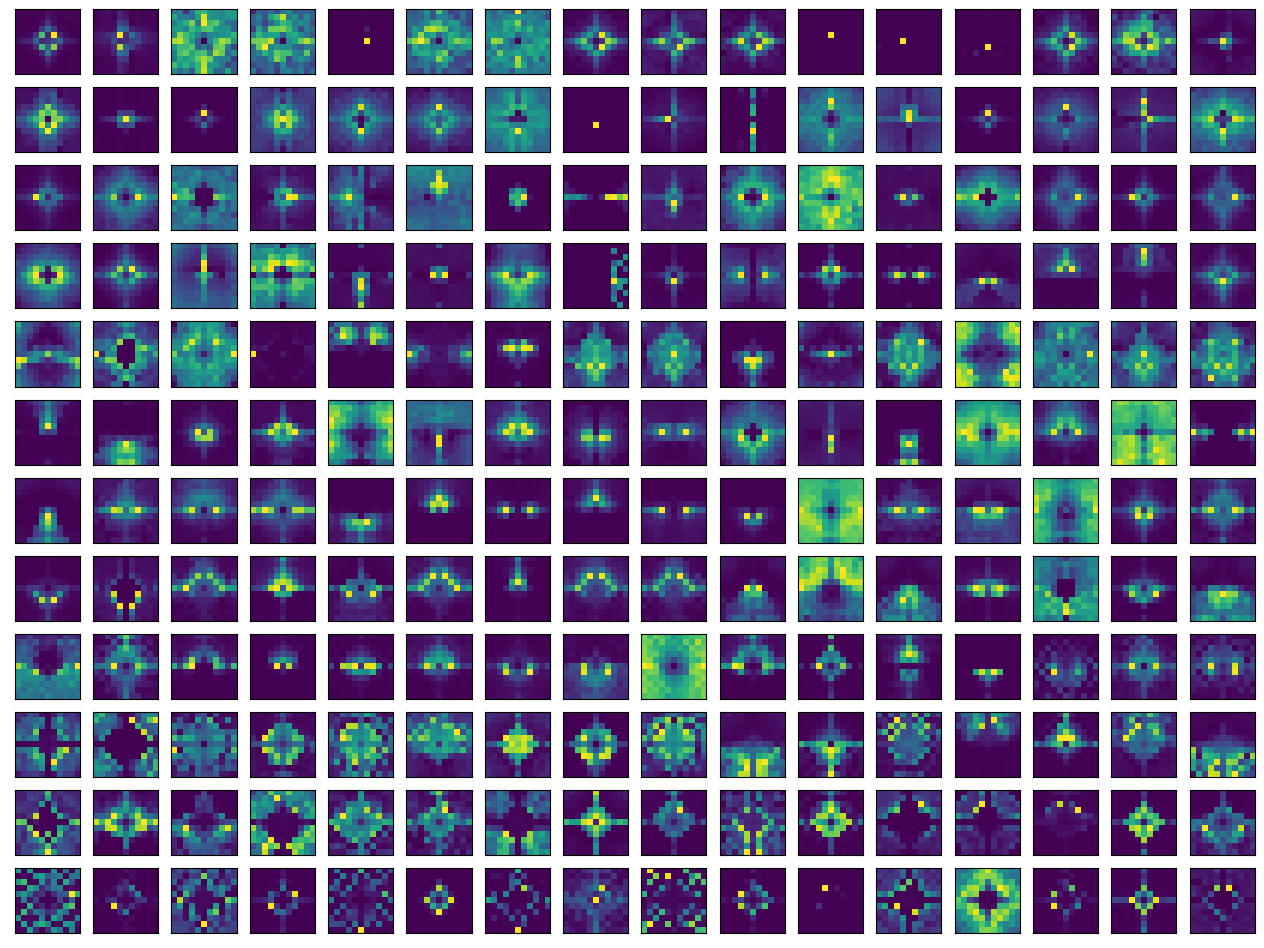}
    \caption{Exponentiated position biases learned by AFT-conv, trained on ImageNet-1K. Each row corresponds to a layer, each column corresponds to a head (the first 16 are shown). This model has top 1 accuracy of $80.8\%$.}
    \label{fig:conv_aft_att_full}
\end{figure}

\begin{figure}
    \centering
    \includegraphics[scale=0.4]{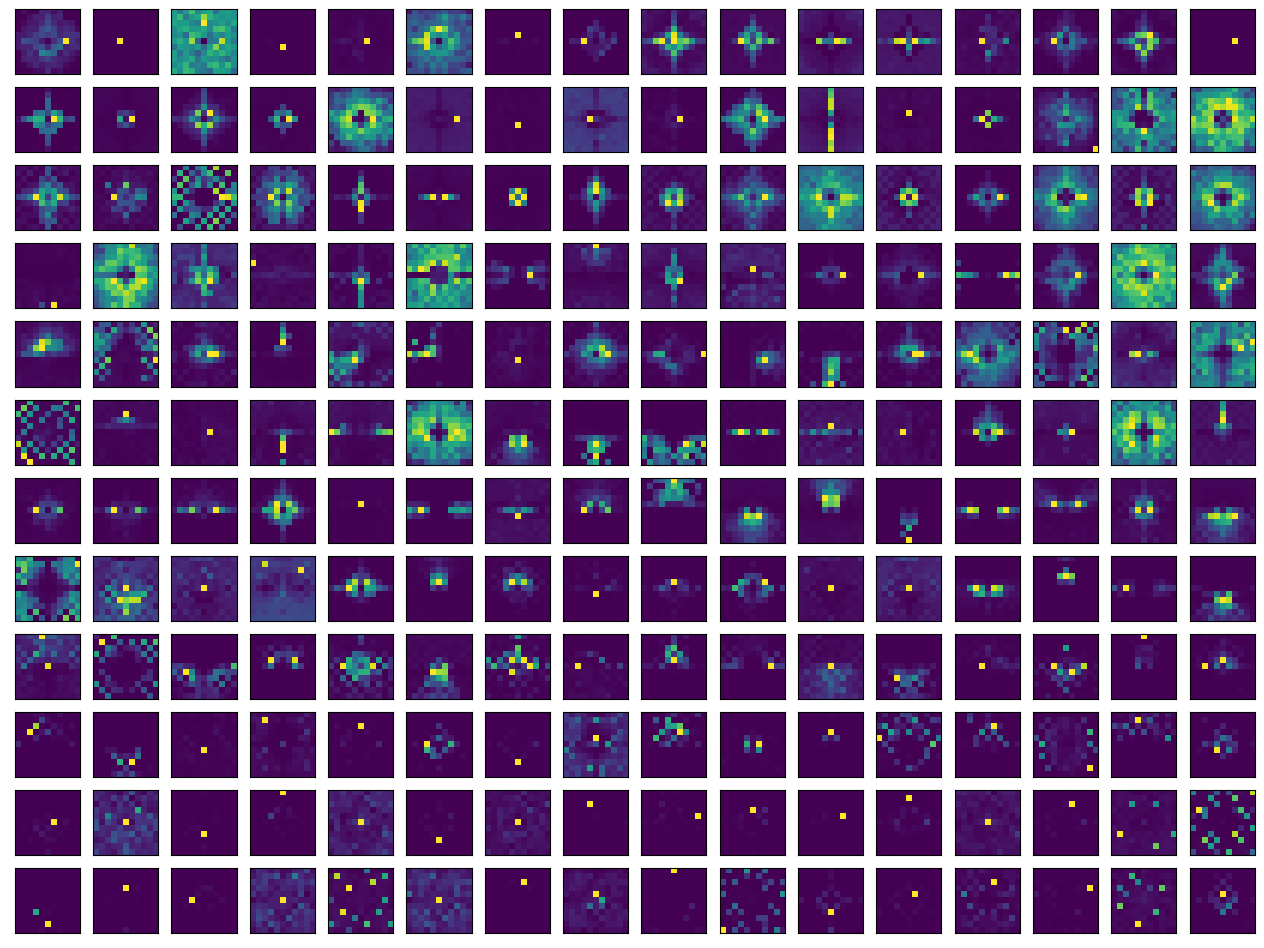}
    \caption{Exponentiated position biases learned by AFT-conv (kernel size $11\times 11$) with \textbf{sparsity regularization}, trained on ImageNet-1K. Each row corresponds to a layer, each column corresponds to a head (the first 16 are shown). This model has top 1 accuracy of $80.9\%$.}
    \label{fig:ent-reg}
\end{figure}

\begin{figure}[t]
    \centering
    \includegraphics[scale=0.4]{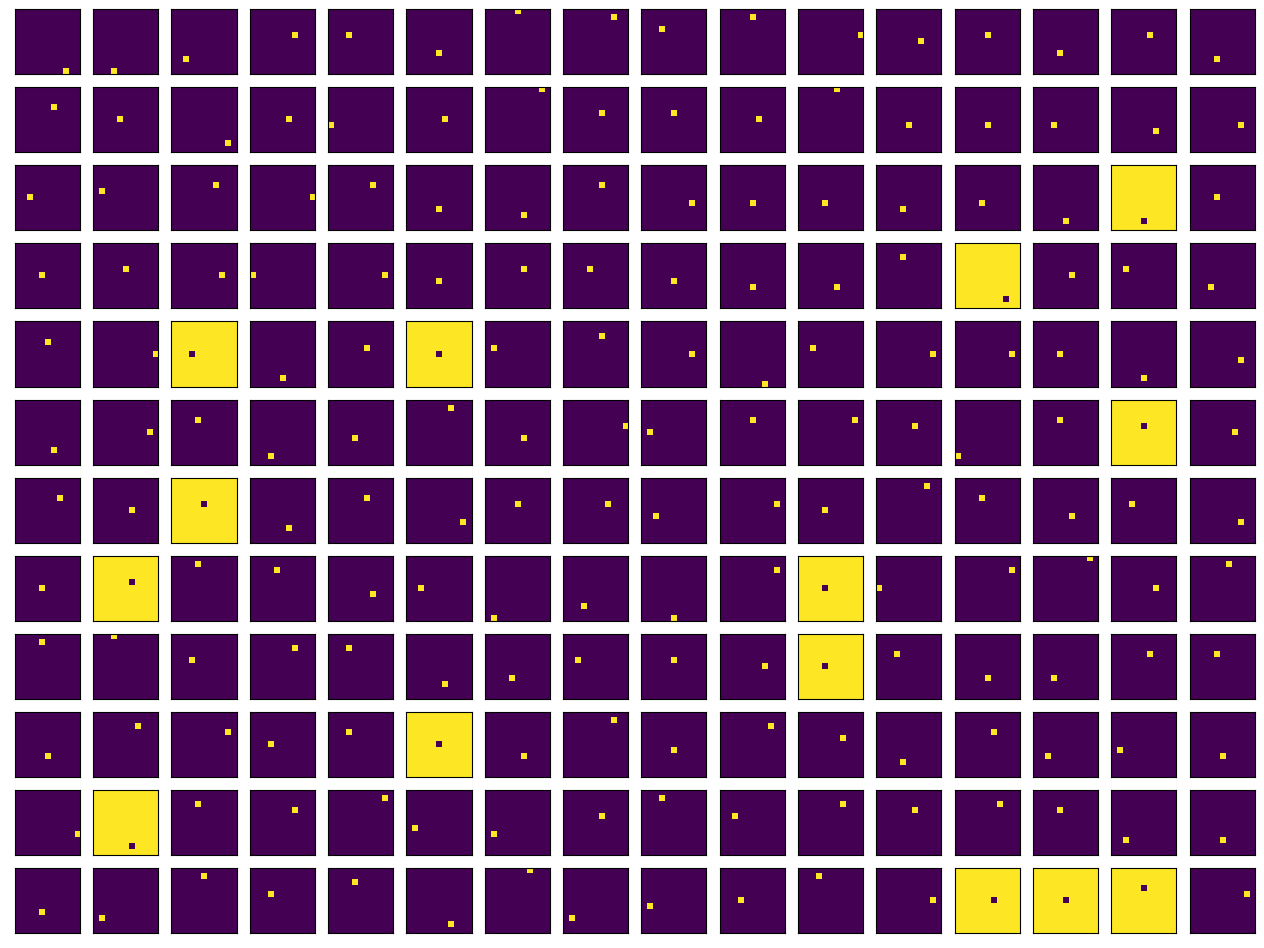}
    \caption{Exponentiated position biases learned AFT-conv (kernel size $11\times 11$) with \textbf{Gumbel softmax sampling}, trained on ImageNet-1K. Each row corresponds to a layer, each column corresponds to a head (the first 16 are shown). This model has top 1 accuracy of $79.9\%$.}
    \label{fig:gumbel}
\end{figure}

\section{Sparsity}
The position biases learned by AFT-conv (kernel size $11\times 11$) as shown in Figure \ref{fig:conv_aft_att_full} demonstrates interesting sparsity patterns, which suggests great potential for quantization and pruning. To this end, we experimented with a simple sparsity promoting regularization term:
\begin{equation}
    reg(w) = \sum_{i=1}^h H(w^i), \; H(w^i) = \text{entropy}(\text{softmax}(w^i)).
\end{equation}
Where we simply minimize the entropy for each head, with the softmax distribution using $w_i$ as the logits. We combining $reg(w)$ with the cross entropy loss with a small weighting ($0.001$) and train with the AFT-conv with kernel size 11 and $384$ heads. This results in a slight improvement in accuracy (due to its regularization effect) of $80.9$ vs $80.8$, as well as sparser looking position biases. The visualization is shown in Fig. \ref{fig:ent-reg}. We see that the position biases are much more sparsely distributed as expected. 

Encouraged by this, we continued to push the sparsity to an extreme form. Now for each head, we only assign a learned relative position bias for a \textbf{single position}. To do this, during training, we multiply the position biases $w$ for each layer and each head with a sample from its corresponding Gumbel softmax distribution \citep{gumbel}:
\begin{equation}
    w^i = w^i * \text{gumbel}(w^i; \tau),
\end{equation}
where $\tau$ is the temperature term for Gumbel softmax, and we set it as 0.5; $\text{gumbel}(w^i; \tau)$ produces a (sparse) sample with the same shape as $w_i$. During inference, the Gumbel softmax is replaced with hard max, i.e., a one hot vector is returned. This results in a model with top 1 accuracy $79.9$, with less than 1 point drop compared with the unregularized model. The position biases are visualized in Fig. \ref{fig:gumbel}. This extreme model variant makes it possible to implement the context reduction of $K, V$ with a combination of global average pooling and indexing, which has the same complexity as AFT-simple but maintains strong performance (comparable to that of the standard Transformer).

\end{document}